\documentclass[pre,twocolumn,amsmath,amssymb,superscriptaddress,letterpaper]{revtex4}        

\usepackage{graphicx}
\usepackage{bm}
\usepackage{subfigure}
\usepackage{url}
\usepackage{epstopdf}

\begin{document}

\title{Fast, simple and accurate handwritten digit classification by training shallow neural network classifiers  with the `extreme learning machine' algorithm}

\author{Mark D. McDonnell}
 \email{mark.mcdonnell@unisa.edu.au}
\affiliation{Computational and Theoretical Neuroscience Laboratory, Institute for Telecommunications Research, University of South Australia, SA 5095, Australia}
\author{Migel D. Tissera}
\affiliation{Computational and Theoretical Neuroscience Laboratory, Institute for Telecommunications Research, University of South Australia, SA 5095, Australia}
\author{Tony Vladusich}
\affiliation{Computational and Theoretical Neuroscience Laboratory, Institute for Telecommunications Research, University of South Australia, SA 5095, Australia}
\author{Andr{\'e} van Schaik}
\affiliation{Biomedical Engineering and Neuroscience Group, The MARCS Institute, The University of Western Sydney, Australia}
\author{Jonathan Tapson}
 \email{J.Tapson@uws.edu.au}
\affiliation{Biomedical Engineering and Neuroscience Group, The MARCS Institute, The University of Western Sydney, Australia}

\date{\today}

\begin{abstract}
Recent advances in training deep (multi-layer) architectures have inspired a renaissance in neural network use. For example, deep convolutional networks are becoming the default option for difficult tasks on large datasets, such as image and speech recognition. However, here we show that error rates below 1\% on the MNIST handwritten digit benchmark can be replicated with shallow non-convolutional neural networks. This is achieved by training such networks using the `Extreme Learning Machine' (ELM) approach, which also enables a very rapid training time ($\sim$10 minutes). Adding distortions,  as is common practise for MNIST, reduces error rates even further. Our methods are also shown to be capable of achieving less than 5.5\% error rates on the NORB image database. To achieve these results, we introduce several enhancements to the standard ELM algorithm, which individually and in combination can significantly improve performance. The main innovation is to ensure each hidden-unit operates only on a randomly sized and positioned patch of each image. This form of random `receptive field' sampling of the input ensures the input weight matrix is sparse, with about 90\% of weights equal to zero. Furthermore, combining our methods with a small number of iterations of a single-batch backpropagation method can significantly reduce the number of hidden-units required to achieve a particular performance. Our close to state-of-the-art results for MNIST and NORB suggest that the ease of use and accuracy of the ELM algorithm for designing a single-hidden-layer neural network classifier should cause it to be given greater consideration either as a standalone method for simpler problems, or as the final classification stage in deep neural networks applied to more difficult problems.
\end{abstract}

\maketitle

\section{Introduction}

The current renaissance in the field of neural networks is a direct result of the success of various types of deep network in tackling difficult classification and regression problems on large datasets.  It may be said to have been initiated by the development of Convolutional Neural Networks (CNN) by LeCun and colleagues in the late 1990s~\cite{LeCun.98} and to have been given enormous impetus by the work of Hinton and colleagues on Deep Belief Networks (DBN) during the last decade~\cite{Hinton.06}.  It would be reasonable to say that deep networks are now considered to be a default option for machine learning on large datasets.

The initial excitement over CNN and DBN methods was triggered by their success on the MNIST handwritten digit recognition problem~\cite{LeCun.98}, which was for several years the standard benchmark problem for hard, large dataset machine learning.  A high accuracy on MNIST is regarded as a basic requirement for credibility in a classification algorithm.  Both CNN and DBN methods were notable, when first published, for posting the best results up to that respective time on the MNIST problem.

{The standardised MNIST database consists of 70,000 images, each of size 28 by 28 greyscale pixels~\cite{MNIST}. There is a standard set of 60,000 training images and a standard set of 10,000 test images, and numerous papers report results of new algorithms applied to these 10,000 test images, e.g.~\cite{LeCun.98,Ciresan.10,Ciresan.12,Wan.13,Zeiler.13,Goodfellow.13,Lee.14}.}

{In this report, we introduce variations of the Extreme Learning Machine algorithm~\cite{Huang.06} and report their performance on the MNIST test set.}  These results are equivalent or superior to the original results achieved by CNN and DBN on this problem, and are achieved with significantly lower network and training complexity.  {This poses the important question as to whether the ELM training algorithm should be a more popular choice for this type of problem, and a more commonplace algorithm as a first step in machine learning.}

Table~\ref{Table1} summarises our results, and shows some comparison points with results obtained by other methods in the past {(note that only previous results that do not use data augmentation methods are shown, and only one of our new results is for such a case). Our new results surpass results using earlier deep networks, but recent regularisation methods such as drop connect~\cite{Wan.13}, stochastic pooling~\cite{Zeiler.13}, dropout~\cite{Goodfellow.13} and so-called `deeply supervised networks'~\cite{Lee.14} have enabled deep convolutional networks to set new state-of-the-art performance for MNIST for the case where no  data-augmentation is used. Nevertheless, our best result for a much simpler single-hidden-layer neural network classifier trained using the very fast ELM algorithm, and without using data augmentation, is within just 41 errors out of 10000 test images of the state-of-the-art.}

\subsection{The Extreme Learning Machine: Notation and Training}

The Extreme Learning Machine (ELM) training algorithm~\cite{Huang.06} is relevant for a single hidden layer feedforward network (SLFN), {similar to a standard neural network. However, there are three key departures from conventional SLFNs.  These are (i) that the hidden layer is frequently very much larger than a neural network trained using backpropagation; 
(ii) the weights from the input to the hidden layer neurons are randomly initialised and are fixed thereafter (i.e., they are not trained); and (iii)} the output neurons are linear rather than sigmoidal in response, allowing the output weights to be solved by least squares regression.   These attributes have also  been combined in learning systems several times previously~\cite{Schmidt.92,Chen.96,Eliasmith.99,Eliasmith}.

The standard ELM algorithm can provide very good results in machine learning problems requiring classification or regression (function optimization); in this paper we demonstrate that it provides an accuracy on the MNIST problem superior to prior reported results for similarly-sized SLFN networks~\cite{LeCun.98,Tapson.14}.

We begin by introducing three parameters that define the dimensions of an ELM used as an $N$-category classifier: $L$ is the dimension of input vectors, $M$ is the number of hidden layer units, and $N$ is the number of distinct labels for training samples. For the case of classifying $P$ test vectors,  it is convenient to define the following matrices:
\begin{itemize}\setlength{\itemsep}{-0.7mm}
\item ${\bf X_{\rm test}}$, of size $L\times P$, is formed by setting each column to equal a single test vector.
\item ${\bf Y}_{\rm test}$, of size $N\times P$, numerically represents the prediction vector of the classifier.
\end{itemize}\vspace{-1em} 
~\\
To map from input vectors to network outputs, two weights matrices are required:
\begin{itemize}\setlength{\itemsep}{-0.7mm}
\item ${\bf W_{\rm in}}$, of size  $M\times L$, contains the input weight matrix that maps length-$L$ input vectors to length $M$ hidden-unit inputs.
\item ${\bf W_{\rm out}}$, of size $N\times M$,  contains the output weights that project from the $M$ hidden-unit activations to a length $N$ class prediction vector.
\end{itemize}\vspace{-1em} 
~\\
We also introduce matrices to describe inputs and outputs to/from the hidden-units:
\begin{itemize}\setlength{\itemsep}{-0.7mm}
\item ${\bf D}_{\rm test}:= {\bf W_{\rm in}}{\bf X_{\rm test}}$, of size $M\times P$, contains the linear projections of the input vectors that are inputs to each of the $M$ hidden-units. A bias for each hidden unit can be added by expanding the size of the input dimension from $L$ to $L+1$, and setting the additional input element to always be unity for all training and test data, with the bias values  included as an additional column in ${\bf W}_{\rm in}$. 
\item ${\bf A}_{\rm test}$, of size $M\times P$, contains the hidden-unit activations that occur due to each training vector, and is given by
\end{itemize}\vspace{-1em} 
\begin{align}
{\bf A}_{\rm test} := f({\bf D}_{\rm test}),
\end{align}
where $f(\cdot)$ is shorthand notation for the fact that each element of ${\bf D}_{\rm test}$ is nonlinearly converted term-by-term to the corresponding element of ${\bf A}_{\rm test}$. For example, if the hidden unit response is given by the logistic sigmoid function, then 
\begin{align}
({\bf A}_{\rm test})_{i,j} &= f(({\bf D}_{\rm test})_{i,j})=\frac{1}{1+\exp{(-({\bf D}_{\rm test})_{i,j})}}.\label{logistic}
\end{align}
Many  nonlinear activation functions can be equally effective, such as the {rectified linear unit (ReLU) function~\cite{Nair.10},} the absolute value function or the quadratic function. As with standard artificial neural networks, the utility of the nonlinearity is that it introduces hidden-unit responses that represent correlations or `interactions' between input elements, rather than simple linear combinations of them.

The overall conversion of test data to prediction vectors can be written as
\begin{align}
{\bf Y}_{\rm test}
&= {\bf W}_{\rm out}f({\bf W}_{\rm in}{\bf X}_{\rm test}).
\end{align}

We now describe the ELM training algorithm. We introduce  $K$ to denote the number of training vectors available. It is convenient to introduce the following matrices that are relevant for training an ELM: {${\bf X}_{\rm train}$, of size $L\times K$,  ${\bf A}_{\rm train}$, of size $M\times K$, and ${\bf Y}_{\rm train}={\bf W}_{\rm out}{\bf A}_{\rm train}$, of size $N\times K$ are defined analogously to ${\bf X}_{\rm test}$, ${\bf A}_{\rm test}$ and ${\bf Y}_{\rm test}$ above. We also introduce ${\bf Y}_{\rm label}$, of size $N\times K$, which numerically represents the labels of each class of each training vector; it is convenient to define this mathematically such that each column has a $1$ in a single row, and all other entries are zero. The only 1 entry in each column occurs in the row corresponding to the label class for each training vector.}

Ideally we seek to find to find ${\bf W_{\rm out}}$ that satisfies
\begin{align}\label{T}
{\bf Y}_{\rm label}={\bf W}_{\rm out}{\bf A}_{\rm train}.
\end{align}
However,  the number of unknown variables in ${\bf W}_{\rm out}$ is $NM$, and the number of equations is $NK$. Although an exact solution  potentially exists if $M=K$, it is usually  the case that $M< K$ (i.e., there are many more training samples than hidden units) so that the system is overcomplete. The usual approach then, is to seek the solution that minimises the mean square error between ${\bf {Y}_{\rm label}}$ and ${\bf {Y}_{\rm train}}$. { This is a standard least squares regression problem for which the exact solution is ${\bf W}_{\rm out} = {\bf Y}_{\rm label}{\bf A}_{\rm train}^{\top}({\bf A}_{\rm train}{\bf A}_{\rm train}^{\top})^{-1}$}, assuming that the inverse exists (in practice it usually does).

{It can also be useful to regularise the problem to reduce overfitting, by ensuring that the weights of ${\bf W}_{\rm out}$ do not become large. The standard ridge-regression approach~\cite{Marquardt.75} produces the following closed form solution for the output weights:}
\begin{align}\label{Q}
{\bf W}_{\rm out} &= {\bf Y}_{\rm label}{\bf A}_{\rm train}^{\top}({\bf A}_{\rm train}{\bf A}_{\rm train}^{\top}+c{\bf I})^{-1},
\end{align}
where ${\bf I}$ is the $M\times M$ identity matrix, and $c$ can be optimised using cross-validation techniques. {As is discussed in more detail below, we have found  QR decomposition~\cite{Press} to be the most effective method for  solving for ${\bf W}_{\rm out}$.}

\section{Faster and more accurate performance by shaping the input weights non-randomly}
 
 In the conventional ELM algorithm, the input weights are randomly chosen, typically from a continuous uniform distribution on the {interval $[-1,1]$~\cite{Huang.12}, but we have found that other distributions such as bipolar binary values from $\{-1,1\}$ are equally effective.}
 
 Beyond such simple randomisation of the input weights, small improvements can be made by ensuring the rows of ${\bf W}_{\rm in}$ are as mutually orthogonal as possible~\cite{Kasun.13}. This cannot be achieved exactly unless $M \le L$, but simple random weights typically produce dot products of distinct rows of ${\bf W}_{\rm in}$ that are close to zero, albeit not exactly zero, while a dot product of each row with itself is always much larger than zero.  In addition, it can be beneficial to normalise the length of each row of ${\bf W}_{\rm in}$, as occurs in the orthogonal case.

In contrast, we can also aim to find weights that, rather than being selected from a random distribution, are instead chosen to be well {\em matched} to the statistics of the data, with the hope that this will improve generalisation of the classifier. Ideally we do not want to have to learn these weights, but rather just form the weights as a simple function of the data.

Here we focus primarily on improving the performance of the ELM algorithm by biasing the selection of input layer weights in six different ways, several of which were recently introduced in the literature, and several of which are novel in this paper.  These methods are as follows: 
\vspace{-0.5em} \begin{enumerate}\setlength{\itemsep}{-0.7mm}
\item Select input layer weights that are random, but biased using the training data samples, so that the dot product between weights and training data samples is likely to be large.  This is called Computed Input Weights ELM (CIW-ELM)~\cite{Tapson.14}.  
\item Ensure input weights are constrained to a set of difference vectors of between-class samples in the training data.  This is called Constrained ELM (C-ELM)~\cite{Zhu.15}.  
\item Restrict the weights for each hidden layer neuron to be non-zero only for a small, random rectangular patch of the input visual field; we call this Receptive Field ELM (RF-ELM).  Although we believe this method to be new to ELM approaches, it is inspired by other machine learning approaches that aim to mimic cortical neurons that have limited visual receptive fields, such as convolutional neural networks~\cite{Coates.11}.
\item Combine RF-ELM with CIW-ELM, or RF-ELM with C-ELM; we show below that the combination is superior to any of the three methods individually.
\item Pass the results of a RF-CIW-ELM and a RF-C-ELM into a third standard ELM (thus producing a two-layer ELM system); we show below that this gives the best overall performance of all methods considered in this paper.
\item Application of the backpropagation method of~\cite{Yu.12}. This method highlights that the performance of an ELM can be enhanced by adjusting all input layer weights simultaneously, based on all training data. The output layer weights are maintained in their least-squares optimal state by recalculating them after input layer backpropagation updates. The process of backpropagation updating of the input weights followed by standard ELM recalculation of the output weights can be repeated iteratively until convergence.
\end{enumerate}

RF-ELM and its combination with CIW-ELM and C-ELM, and the two-layer ELM are reported here for the first time. We demonstrate below that each of these methods independently improves the performance of the basic ELM algorithm, and in combination they produce results equivalent to {many} deep networks on the MNIST problem~\cite{LeCun.98,Hinton.06}. First, however, we now describe each method in detail.

 
\subsection{Computed Input Weights for ELM}

\begin{figure}[h]
  \includegraphics[width=1\textwidth]{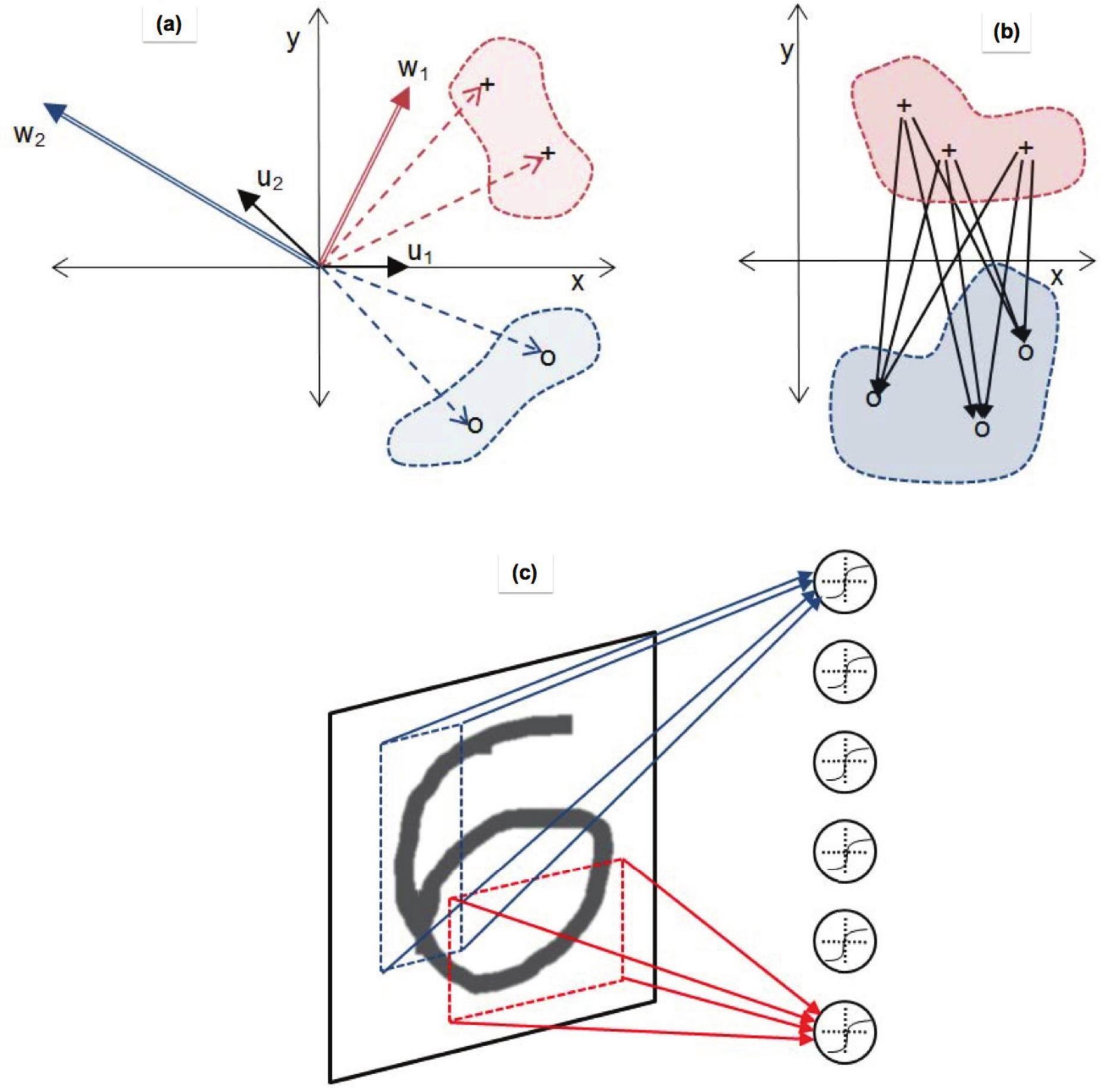}
\caption{{\bf Illustration of the three core methods of shaping ELM input weights.}  In (a), which is a cartoon of the Computed Input Weights ELM (CIW-ELM) process~\cite{Tapson.14}, two classes of input data are indicated by `+' and `o' symbols.  The vectors to the `+' symbols are multiplied by random bipolar binary $\{-1,1\}$) vectors ${\bf u}_{\rm 1}$ and  ${\bf u}_{\rm 2}$ to produce a biased random weight vector  ${\bf w}_{\rm 1}$.  Similarly the weights to the `o' class are also multiplied by random vectors  ${\bf u}_{\rm 1}$ and  ${\bf u}_{\rm 2}$ to produce a biased random weight vector  ${\bf w}_{\rm 2}$.  Note that in practice we would not use the same random binary vectors. In  (b), we show the Constrained ELM (C-ELM) process~\cite{Zhu.15}.  The black arrows are weight vectors derived by computing the difference of two classes; in this case, the difference between the `+' elements and the `o' elements.  In (c), we illustrate the Receptive Field ELM (RF-ELM) method; weights for each hidden layer neuron are restriced to being non-zero for only a small random rectangular receptive field in the original image plane.}\label{Fig1}
\end{figure}

The CIW-ELM approach is motivated by considering the standard backpropagation algorithm~\cite{Rumelhart.86}.  A feature of weight-learning algorithms is that they operate by adding to the weights some proportion of the training samples, or a linear sum or difference of training samples.  In other words, apart from a possible random initialization, the weights are constrained to take final values which are drawn from a space defined in terms of linear combinations of the input training data as basis vectors---see Fig.~\ref{Fig1}.  While it has been argued, not without reason, that it is a strength of ELM that it is not thusly constrained~\cite{Huang.14}, the use of this basis as a constraint on input weights will bias the ELM network towards a conventional (backpropagation) solution.

The CIW-ELM algorithm is as follows~\cite{Tapson.14}:
\vspace{-0.5em} \begin{enumerate}\setlength{\itemsep}{-0.7mm}
\item	For use in the following steps only, normalize all training data by subtracting the mean over all training points and dimensions and then dividing by the standard deviation.
\item	Divide the $M$ hidden layer neurons into $N$ blocks, one for each of $N$ output classes; for data sets where the number of training data samples for each class are equal, the block size is  $M_n = M/N$.  We denote the number of training samples per class as $K_n,~n=1,\dots,N$.  If the training data sets for each class are not of equal size, the block size can be adjusted to be proportional to the data set size.
\item	For each block, generate a random sign ($\pm 1$) matrix, ${\bf R}_n$ of size $M_n \times K_n$.
\item	Multiply ${\bf R}_n$ by the transpose of the input training data set for that class, $X_{{\rm train},n}^{\top}$, to produce $ M_n \times L$ summed inner products, which are the weights for that block of hidden units.
\item	Concatenate these $N$ blocks of weights for each class into the $M\times L$ input weight matrix ${\bf W_{\rm in}}$.
\item	Normalize each row of the input weight matrix, ${\bf W_{\rm in}}$, to unity length.
\item	Solve for the output weights of the ELM using standard ELM methods described above.
\end{enumerate}\vspace{-1em} 

\subsection{Constrained Weights for ELM}

Recently, Zhu {\em et al.}~\cite{Zhu.15} have published a method for constraining the input weights of ELM to the set of difference vectors of between-class samples.  The difference vectors of between-class samples are the set of vectors connecting samples of one class with samples of a different class, in the sample space---see Fig.~\ref{Fig1}.  In addition,  a methodology is proposed for eliminating from this set the vectors of potentially overlapping spaces (effectively, the shorter vectors) and for reducing the use of near-parallel vectors, in order to more uniformly sample the weight space.

The Constrained ELM (C-ELM) algorithm we used is as follows~\cite{Zhu.15}:
\vspace{-0.5em} \begin{enumerate}\setlength{\itemsep}{-0.7mm}
\item	Randomly select $M$ distinct pairs of training data such that:
\vspace{-0.5em} \begin{enumerate}\setlength{\itemsep}{-0.7mm}
\item  each pair comes from two distinct classes;
\item   the vector length of the difference between the pairs is smaller than some constant, $\epsilon$.
\end{enumerate}
\item Set each row of the $M\times L$ input weight matrix ${\bf W_{\rm in}}$ to be equal to the difference between each pair of randomly selected training data.
\item Set the bias for each hidden unit equal to the scalar product of the sum of each pair of randomly selected training data and the difference of each pair of randomly selected training data.
\item	Normalize each row of the input weight matrix, ${\bf W_{\rm in}}$, and each bias value, by the vector of the difference of the corresponding pair of randomly selected training data. 
\item	Solve for the output weights of the ELM using standard ELM methods described above.
\end{enumerate}\vspace{-1em} 

\subsection{Receptive Fields for ELM}

We have found that a data-blind (unsupervised) manipulation of the input weights improves generalization performance. The  approach has the added bonus that the input weight matrix is sparse, with a very high percentage of zero entries, which could be advantageous for hardware implementations, or if sparse matrix storage methods are used in software. 

The RF-ELM approach is inspired by neurobiology, and strongly resembles some other machine learning {approaches~\cite{Coates.11}}. Biological sensory neurons tend to be tuned with preferred receptive fields so that they receive input only from a subset of the overall input space. The region of responsiveness tends to be contiguous in some pertinent dimension, such as space for the visual and touch systems, and frequency for the auditory system. Interestingly, this contiguity aspect may be lost beyond the earliest neural layers, if features are combined  randomly.

In order to loosely mimic this organisation of biological sensory systems, in this paper where we consider only  image classification tasks, for each hidden unit we create randomly positioned and sized rectangular masks that are smaller than the overall image. These masks ensure only a small subset of the length-$L$ input data vectors influence any given hidden unit---see Fig.~\ref{Fig1}.

The algorithm for generating these `receptive-field' masks is as follows:
\vspace{-0.5em} \begin{enumerate}\setlength{\itemsep}{-0.7mm}
\item Generate a random input weight matrix ${\bf W}$ (or instead start with a CIW-ELM or C-ELM input weight matrix).
\item For each of $M$ hidden units, select two  pairs of distinct random integers from $\{1,2,\dots L\}$ to form the coordinates of a rectangular mask.
\item If any mask has total area smaller than some value $q$ then discard and repeat.
\item Set the entries of a $\sqrt{L}\times \sqrt{L}$ square matrix that are defined by the two pairs of integers to $1$, and all other entries to zero.
\item Flatten each receptive field matrix into a length $L$ vector where each entry corresponds to the same pixel as the entry in the data vectors ${\bf X}_{\rm test}$ or ${\bf X}_{\rm train}$.
\item Concatenate the resulting $M$ vectors into a receptive field matrix ${\bf F}$ of size $M\times L$.
\item Generate the ELM input weight matrix by finding the Hadamard product (term by term multiplication) ${\bf W}_{\rm in}={\bf F} \circ {\bf W}$.
\item	Normalize each row of the input weight matrix,  ${\bf W_{\rm in}}$, to unity length.
\item	Solve for the output weights of the ELM using standard ELM methods described above.
\end{enumerate}

We have additionally found it beneficial to exclude pixels from the mask if most or all training images have identical values for those regions. For the MNIST database, this typically means ensuring all masks exclude the first and last 3 rows and first and last 3 columns.  For MNIST we have found that a reasonable value of the minimum mask size is $q=10$, which enables masks of size $1\times 10$ and $2 \times 5$ and larger, but not $3 \times 3$ or smaller. 

Note that unlike the random receptive field patches described in~\cite{Coates.11} for use in a convolutional network, our receptive field masks are not of uniform size, and not square; we found it beneficial to ensure  a large range of receptive field areas, and large diversity in ratios of lengths to widths.

\subsection{Combining RF-ELM with CIW-ELM and C-ELM}

All three approaches described so far provide weightings for pixels for each hidden layer unit. CIW-ELM and C-ELM weight the pixels to bias hidden-units towards a larger response for training data from a specific class. The sparse weightings provided by RF-ELM bias hidden-units to respond to pixels from specific parts of the image.

We have found that enhanced classification performance can be achieved by combining the shaped weights obtained by either CIW-ELM or C-ELM with the receptive field masks provided by RF-ELM.  The algorithm for either RF-CIW-ELM or RF-C-ELM is as follows.

\vspace{-0.5em} \begin{enumerate}\setlength{\itemsep}{-0.7mm}
\item Follow the first 5 steps of the above CIW-ELM or the first 2 steps of the C-ELM algorithm, to obtain an un-normalized shaped input weight matrix, ${\bf W_{{\rm in},s}}$.
\item Follow the first 6 steps of the RF-ELM algorithm to obtain a receptive field matrix,  ${\bf F}$.
\item Generate the ELM input weight matrix by finding the Hadamard product (term by term multiplication) ${\bf W}_{\rm in}={\bf F} \circ {\bf W_{{\rm in},s}}$.
\item	Normalize each row of the input weight matrix,  ${\bf W_{\rm in}}$, to unity length.
\item If RF-C-ELM, produce the biases according to steps 3 and 4 of the C-ELM algorithm, but use the masked difference vectors rather than the unmasked ones.
\item	Solve for the output weights of the ELM using standard ELM methods described above.
\end{enumerate}\vspace{-1em} 

\subsection{Combining RF-C-ELM with RF-CIW-ELM in a two-layer ELM: RF-CIW-C-ELM}

\begin{figure}[h]
\includegraphics[width=1\textwidth]{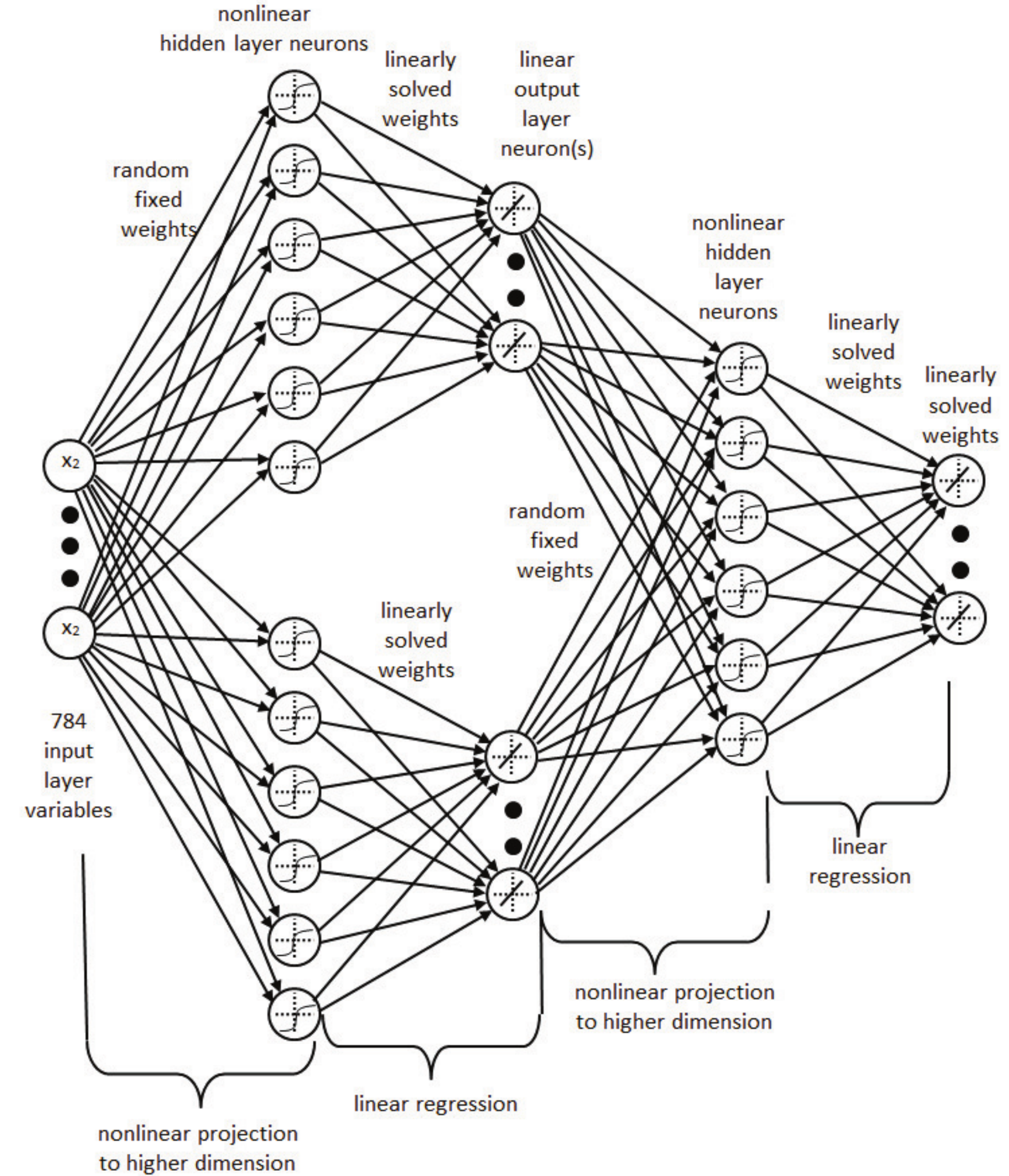}
\caption{{\bf Combined two-layer RF-CIW-ELM and RF-C-ELM network.} This figure depicts the structure of our multilayer ELM network that combines a CIW-RF-ELM network with a C-ELM network, using what is effectively an autoencoder output.  Note that the middle linear layer of neurons can be removed by combining the output layer weights of the first network with the input layer weights of the second; we have not shown this here, in order to clarify the development of the structure.}\label{Fig2}
\end{figure}

We have found that results obtained with RF-C-ELM and RF-CIW-ELM are similar in terms of error percentage when applied to the MNIST benchmark, but the errors follow different patterns.  As such, a combination of the two methods seemed to offer promise.  We have combined the two methods using a multiple-layer ELM which consists of an  RF-C-ELM network and a RF-CIW-ELM network in parallel, as the first two layers.  The outputs of these two networks are then combined using a further ELM network, which can be thought of as an ELM-autoencoder, albeit that it has twenty input neurons and ten output neurons; the input neurons are effectively two sets of the same ten labels.  The structure is shown in Fig.~\ref{Fig2}. The two input networks are first trained to completion in the usual way, then the autoencoder layer is trained using the outputs of the input networks as its input, and the correct labels as the outputs.  The result of this second-layer network, which is very quick to implement (as it uses a hidden layer of typically only 500-1000 neurons), is significantly better than the two input networks (see Table I).  Note that the middlemost layer shown in Fig.~\ref{Fig2} consists of linear neurons, and therefore it can be removed by combining its input and output weights into one connecting weight matrix. However, it is computationally disadvantageous to do so because the number of multiplications will increase.

\subsection{Fine Tuning by Backpropagation}

{In all of the variations of ELM described in this report, the hidden layer weights are not  updated or trained according to the output error, and the output layer weights are solved using least squares regression.  This considerably reduces the trainability of the network, as the number of free parameters is restricted to the output layer weights, which are  generally $\sim10^4$ in number.  It has been argued that any network in which the total number of weights in the output layer is less than the number of training points will likely be enhanced by using backpropagation to train weights in previous layers~\cite{Widrow.13}. Hence,  following} the example of deep networks, and specific ELM versions of backpropagation~\cite{Yu.12}, we have experimented by using backpropagation to fine-tune the hidden layer weights.  This does re-introduce the possibility of overfitting, but that is a well-understood problem in neural networks and the usual methods for avoiding it will apply here.  For simplicity, a batch mode backpropagation was implemented, using the following algorithm. Note that as in Eqn.~(\ref{logistic}), we assume a logistic activation function in the hidden layer neurons, for which the derivative can be expressed as $f' = f(1-f)$.
\vspace{-0.5em} \begin{enumerate}\setlength{\itemsep}{-0.7mm}
\item Construct the ELM and solve for the output layer weights ${\bf W}_{\rm out}$ as described above.
\item Perform iterative backpropagation as follows:
\vspace{-0.5em} \begin{enumerate}\setlength{\itemsep}{-0.7mm}
\item	Compute the error for the whole training set: ${\bf E} = {\bf Y}_{\rm label} - {\bf W}_{\rm out}{\bf A}_{\rm train}$.
\item	Calculate the weights update, as derived by~\cite{Yu.12}: $\Delta{\bf W}_{\rm in} = \xi\left[({\bf W}_{\rm out}^{\top}{\bf E})\circ({\bf A}_{\rm train}-{\bf A}_{\rm train}\circ{\bf A}_{\rm train})\right]{\bf X}_{\rm train}^{\top}$ where $\xi$ is the learning rate, and $\circ$ indicates the Hadamard product (elementwise matrix multiplication).
\item	Update the weights, ${\bf W}_{\rm in} = {\bf W}_{\rm in} - \Delta{\bf W}_{\rm in}$.
\item	Re-calculate ${\bf A}_{\rm train}$ with the new ${\bf W}_{\rm in}$.
\item	Re-solve for ${\bf W}_{\rm out}$ using least squares regression and continue.
\item Repeat from step a) for a desired number of iterations or until convergence.
\end{enumerate}
\end{enumerate}
As illustrated in the Results section, this process has shown a robust improvement on all of the SLFN ELM solutions tested here, provided learning rates which maintained stability were used.

\section{Results and discussion for the MNIST benchmark}

\subsection{SLFN with shaped input-weights}

We trained ELMs using each of the six input-weight shaping methods described above, as well as a standard ELM with binary bipolar ($\{-1,1\}$) random input weights plus row-normalisation. Following the normalisation of rows of the input weight matrix to unity, we multiplied the values of the entire input weight matrix by a factor of 2, for all seven methods, as this scaling was found to be close to optimal in most cases.

\begin{figure}[h]
\includegraphics[width=1\textwidth]{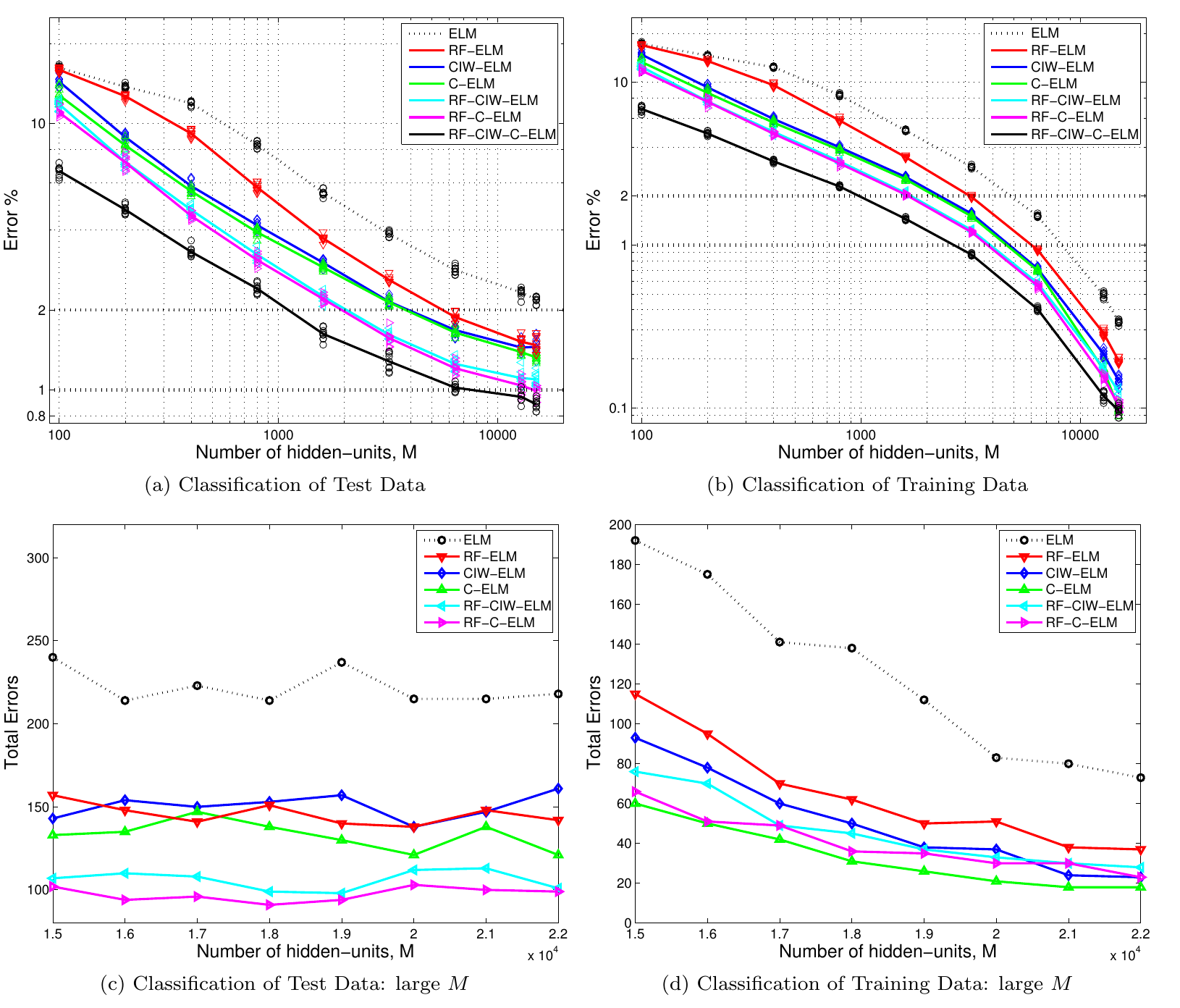}
\caption{{\bf Error rates for MNIST images for various SLFN ELM methods with shaped input weights.} The first row shows the mean  error percentage from 10 different trained networks applied to classify (a) the 10000-point MNIST test data set, and (b) the 60000-point MNIST training data set used to train the networks, for various different sizes of hidden layer, $M$. Markers show the actual error percentage from each of the 10 networks. Note that the data for the combination RF-CIW-C-ELM method is plotted against $M$ used in just one of the three parts of the overall network; the total number of hidden units used is actually $2M+500$. Therefore RF-CIW-C-ELM does not outperform the other methods for the same total number of hidden-units for small $M$. However it can be seen that for large $M$ RF-CIW-C-ELM produces results below 1\% error on the test data set and provide  the best error rates overall.  The second row illustrates that increasing the number of hidden-units above about $M=15000$ leads to overfitting, since as shown in (c), the total number of errors plateaus, whilst  the total number of errors on the training set continues to decrease (shown in (d)). Note that (c) and (d) show results from a single trained network only.}\label{Fig3}
\end{figure}

Our results are shown in Fig.~\ref{Fig3}. To obtain an indication of variance resulting from the randomness inherent in each input-weight shaping method, we trained 10 ELMs using each method, and then plotted the ensemble mean as a function of hidden-layer size, $M$. We also plotted (see markers) the actual error rates for each trained network. It can be seen in Fig.~\ref{Fig3}A that the error rate decreases approximately log-linearly with the log of $M$ for small $M$, before slowing as $M$ approaches about $10^4$. Fig.~\ref{Fig3}B shows the error rate when the actual training data is used. Since our best test results occur when the error rate on the training data is smaller than $0.2$\%, and we found no significant test error improvement for larger $M$ (see Figs~\ref{Fig3}C and~\ref{Fig3}D) we conclude that increasing $M$ further than shown here produces over fitting. This can be verified by cross-validation on the training data.

  \subsection{ELM with shaped input-weights and backpropagation}

\begin{figure}[h]
\includegraphics[width=1\textwidth]{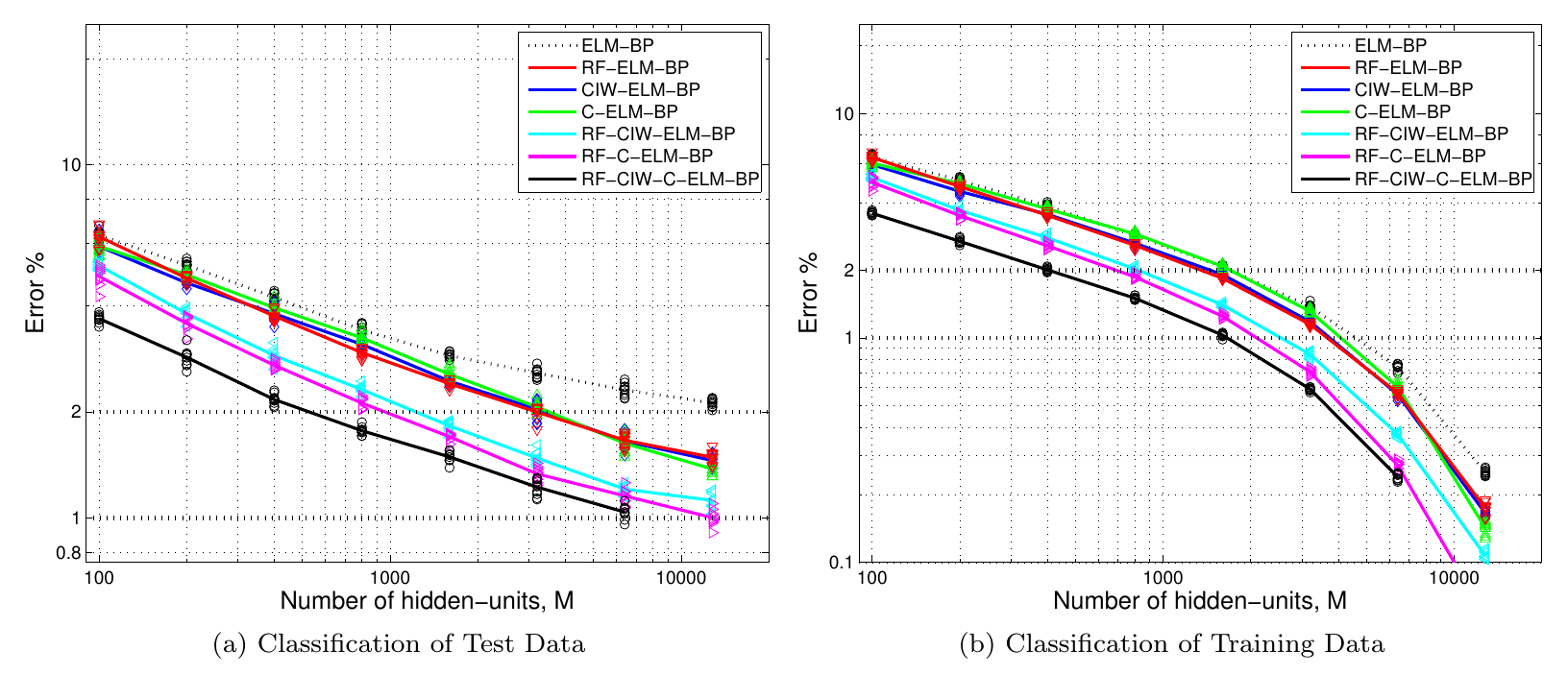}
\caption{{\bf ELM-backpropagation error rates for MNIST  for various SLFN ELM methods with shaped input weights.} Each trace shows the mean  error percentage from 10 different trained networks applied to classify (a) the 10000-point MNIST test data set, and (b) the 60000-point MNIST training data set used to train the networks, for various different sizes of hidden layer, $M$, when ten iterations of backpropagation were also used. Markers show the actual error percentage from each of the 10 networks.  In comparison with Figure~\ref{Fig3}, it can be seen that backpropagation significantly improves the error rate for small $M$ with all methods, but has little impact when $M=12800$.  The total number of hidden units used for RF-CIW-C-ELM is actually $2M+500$, but each parallel ELM has $M$ hidden-units.}\label{Fig4}
\end{figure}

We also trained networks using each of the methods described in the previous section, plus 10 iterations of ELM-backpropagation using a learning rate of $\xi = 0.001$. As can be inferred from the fact that the training set still has relatively high error rates for most $M$ (Figure~\ref{Fig4}B), this use of backpropagation is far from optimal, and does not give  converged  results.  As shown in Figure~\ref{Fig4}A, in comparison with Figure~\ref{Fig3}A, these 10 iterations at a fixed learning rate still provide a significant improvement in the error rate for small $M$.  On the other hand, the improvement for $M=12800$ is minimal, which is not surprising given that the error rate on the training data without using backpropagation for this value of $M$ is already well over 99\% (Figure~\ref{Fig3}B).

It is likely that we can get further enhancements of our error rates by optimising the backpropagation learning rate and increasing the number of iterations used. Moreover, several  methods for accelerating convergence when carrying out  backpropagation have been described previously~\cite{Yu.12}, and we have not used those methods here. However, the best error rate result reported previously for those methods applied to the MNIST benchmark was 1.45\%, achieved with 2048 hidden units, and the best error rate for the backpropagation method we used is 3.73\%~\cite{Yu.12}. Our results for $M\ge 3200$ hidden  units and RF-C-ELM-BP or RF-CIW-ELM-CP match or surpass the former result, despite using the least advanced backpropagation method in prior work~\cite{Yu.12}. The latter figure of 3.73\% is easily surpassed by all 7 methods for just 800 hidden units. Moreover, the training time reported in prior work~\cite{Yu.12} is significantly slower than the time we required when using shaped input weights (see the following section). These outcomes  indicate that the use of input-weight shaping prior to backpropagation  outperforms backpropagation alone by a large margin, in terms of both error rate and training time.


\subsection{Runtime efficiency}

{In many applications, the time required to train a network is not considered as important as the time required for the network to classify new data. However, there do exist applications in which the statistics of training data change rapidly, meaning retraining is required, or deployment of a trained classifier is required very rapidly after data gathering. For example, in financial, sports, or medical data analysis, deployment of a newly trained classifier can be required within minutes of acquiring data, or  retraining may be required periodically, e.g.  hourly.  Hence, we emphasize in this paper the rapidity of training. The speed for testing is negligible in comparison.}

\begin{figure}[h]
\centering
\includegraphics[width=0.7\textwidth]{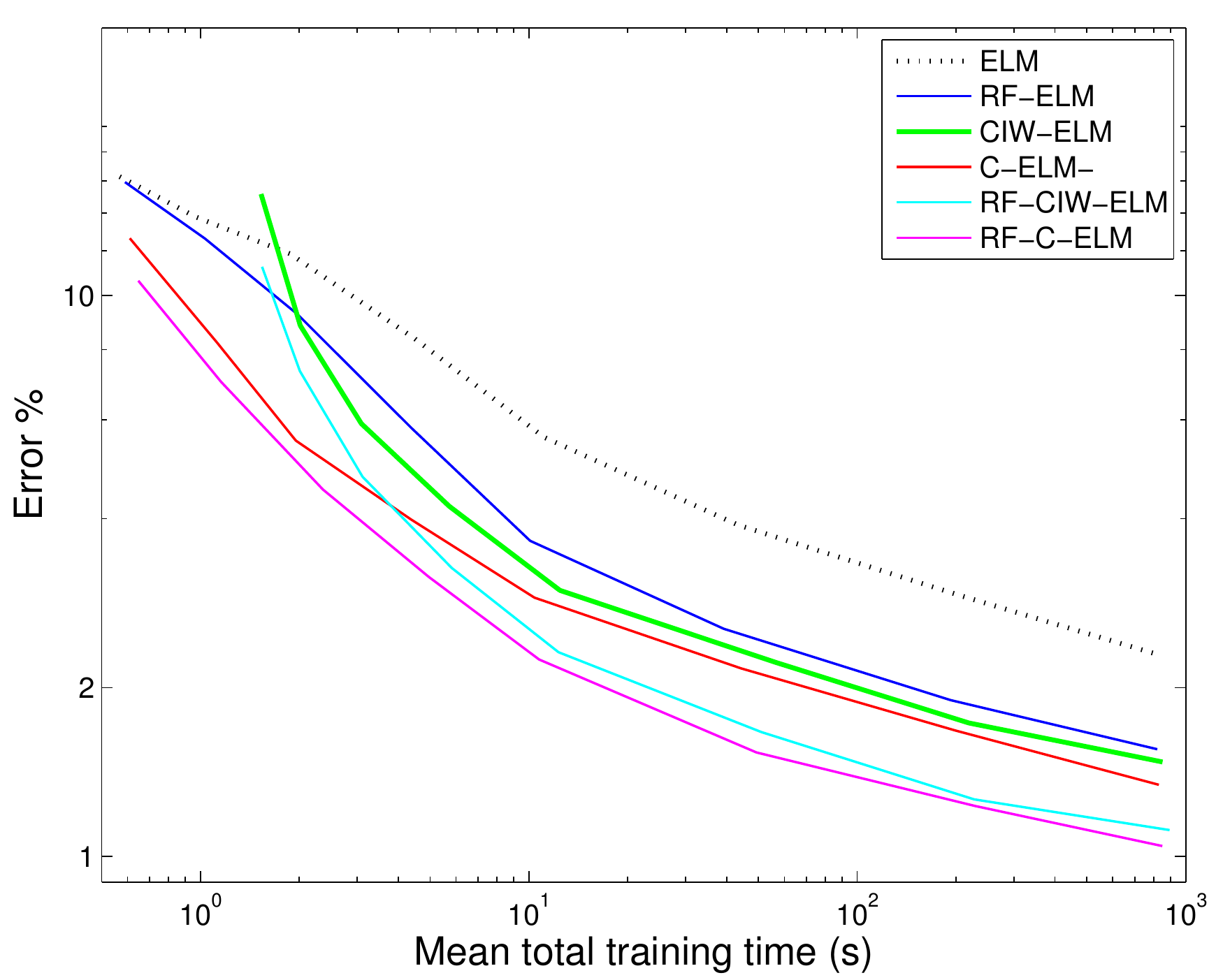}
\caption{{\bf Mean training times for MNIST for various SLFN ELM training methods with shaped input weights.} Each trace shows the mean  run time from 10 different networks trained on all 60000 MNIST training data points, {to achieve the test-date error rates shown in Figure~\ref{Fig4}}. The total time for setup and training  are shown, excluding time to load the MNIST data from files. When backpropagation is applied, the runtime scales approximately linearly with the number of iterations, but each backpropagation iteration is slower than each trace shown here, because both input and output weights are updated in each iteration. {The time for testing is not included in the figure, but was approximately 10 seconds for $M=12,800$, and increases only linearly with $M$.} }\label{Fig5}
\end{figure}

{The mean training runtime} for each of our methods is shown in Figure~\ref{Fig5}.  They were obtained using Matlab running on a Macbook pro with a 3 GHz Intel Core i7 (2 dual cores, for a total of 4 cores), running OS X 10.8.5 with 8 GB of RAM. The times plotted in Figure~\ref{Fig5} are the total times for {setup and training}, excluding time required to load the MNIST data into memory from files. The version of Matlab we used by default  exploits  all four CPU cores for matrix multiplication and least squares regression. Note that the differences in run time for each method are negligible, which is expected, since the most time-consuming part is the formation of the matrix ${\bf A}_{\rm train}{\bf A}_{\rm train}^{\top}$. {The time for testing was not included in the shown data. We found, predictably, that this scaled linearly with $M$, and was about 10 seconds for $M=12,800$.}

The most important conclusion we draw in terms of runtime is that our best results shown here (for $M=15000$ hidden units) for RF-CIW-ELM or RF-C-ELM individually take in the order of 15 minutes total runtime and achieve $\sim$99\% correct classification on MNIST. In comparison, data tabulated previously for backpropagation shows at least 81 minutes in order to achieve 98\% accuracy, and a best result of 98.55\% in 98 minutes~\cite{Yu.12}. In contrast, runtimes reported for the standard ELM algorithm previously (28 seconds for 2048 hidden units~\cite{Yu.12}) are comparable to ours (12 seconds for 1600 hidden units and less than 1 minute for 3200 hidden units). This illustrates that improving error rate by shaping the input weights as we have done here has substantial benefits for runtime and error rate in comparison to backpropagation.

\subsection{Distorting and pre-processing the training set}

{Many other approaches to classifying MNIST handwritten digits improve their error rates by preprocessing it and/or by expanding the size of the  training set by applying affine (translation, rotation and shearing), and elastic distortions of the standard set~\cite{Simard.03,Ciresan.10,Ciresan.12}.   We have also experimented with distorting the training set to improve on the error rates reported here.} For example, with 1 and 2 pixel affine translations, we are able to achieve error rates smaller than 0.8\%. When we added random rotations, scalings, shears and elastic distortions, we achieved a best repeatable error rate of 0.62\%, and an overall best error rate of 0.57\%. However, adding distortions of the training set substantially increases the runtime for two reasons. First, more training points generally requires a larger hidden layer size. For example, when we increase the size of the training set by a factor of 10, we have found we need $M>20000$ to achieve error rates smaller than 0.7\%. This significantly affects the run time through the  $O(M^2)$ matrix multiplication required.

At this stage, we have chosen to not systematically continue to improve the way in which we implement distortions to approach state of the art MNIST results, but our preliminary results show that ELM training is  capable of using such methods to enhance error rate performance, at the expense of a significant increase in runtime, as is expected in other non-ELM methods.

~\\
~\\

\subsection{{Results on NORB}}

{We briefly present some results on a second well-known image classification benchmark: the NORB-small database~\cite{LeCun.04}. This database consists of 48600 stereo greyscale images from five classes, and there is a standard set of 24300 stereo images for training, and a standard set of 24300 for testing. Each of the images in the two stereo channels for each sample is of size 96$\times$96 pixels.}

{Given the large size of each image relative to MNIST images, we preprocessed all images by spatially lowpass filtering using a $9\times9$ pixel Gaussian kernel, with standard deviation of 4, and then decimating to $13\times 13$ pixel images. We then contrast-normalised each image by subtracting its mean, and dividing by its standard deviation.}

\begin{figure}[h]
\centering
\includegraphics[width=0.6\textwidth]{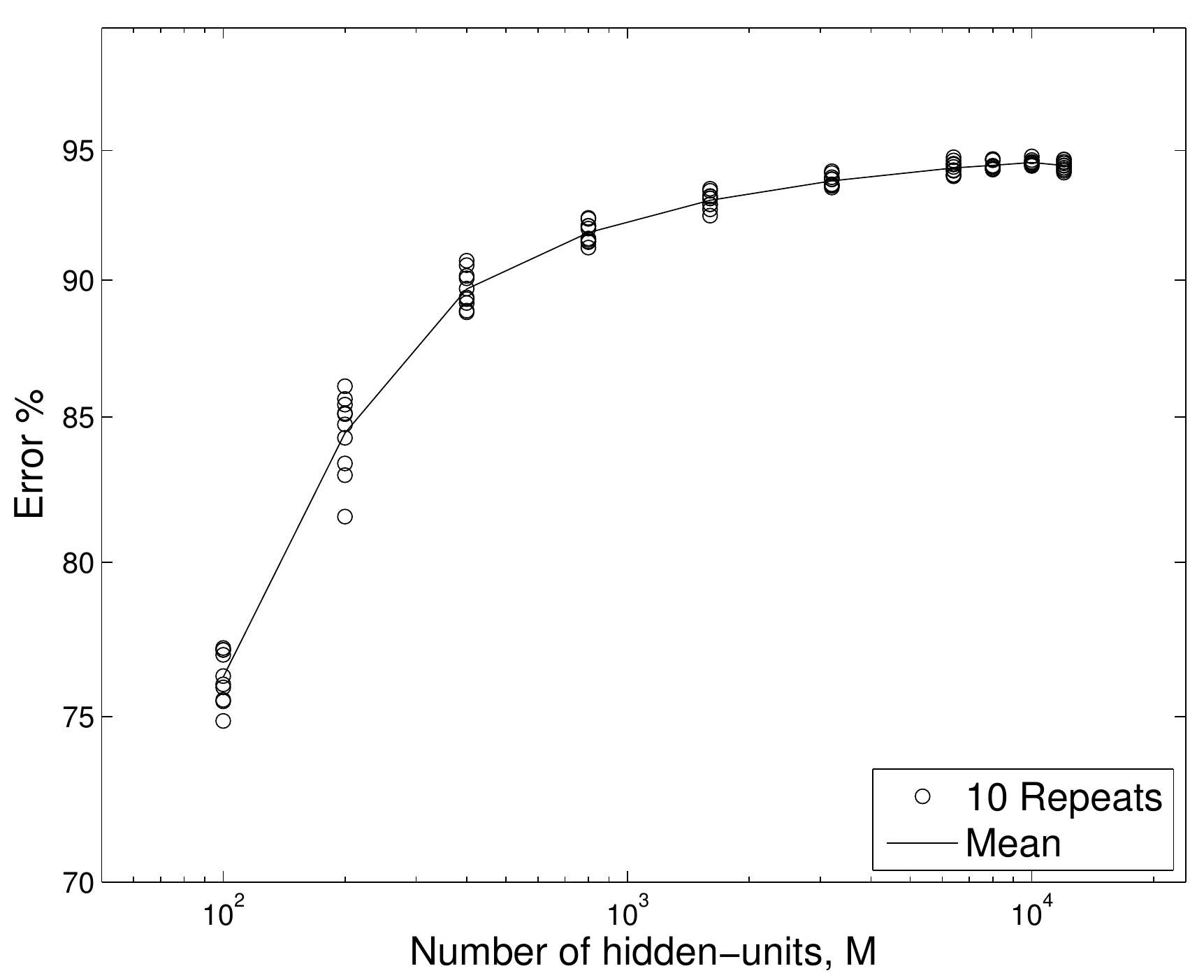}
\caption{{\bf Error rates for NORB-small for RF-C-ELM.} {The error rate on the 24300 stereo-channel NORB-small test images as a function of the number of hidden-units, $M$. The data was preprocessed by downsampling each channel of each image to $13\times 13$ pixels, and then contrast normalising. Our best result from all repeats was $94.76$\%, for $M=10000$.}}\label{Fig6}
\end{figure}

{Figure~\ref{Fig6} shows  results for the error rate (ten repeats and the ensemble mean are shown) on the test set from application of the RF-C-ELM method, as the number of hidden units increases. We set the minimum receptive field size to 1, and the ridge regression parameter to $c=5\times 10^{-6}$. Our test results peak at close to 95\% correct, which is within 3\% of state-of-the-art~\cite{Ciresan.11}, and superior to some results for deep convolutional networks~\cite{Jarrett.09}.}

\subsection{Computationally efficient methods for ELM training: iterative methods for large training sets}

In practice, it is  known to be generally computationally more efficient  (and avoids other potential problems, such as those described in~\cite{Widrow.13a})  to avoid explicit calculation of matrix inverses or pseudo-inverses when solving linear sets of equations. The same principle applies when using the ELM training algorithm, and hence, it is preferable to avoid explicit calculation of the inverse in Eqn.~(\ref{Q}), and instead treat the following as a set of $NM$ linear equations to be solved for $NM$ unknown variables:
\begin{align}\label{Final}
{\bf Y}_{\rm label}{\bf A}_{\rm train}^{\top} &= {\bf W}_{\rm out}({\bf A}_{\rm train}{\bf A}_{\rm train}^{\top}+c{\bf I}).
\end{align}
Fast methods for solving such equations exist, such as the QR decomposition method~\cite{Press}, which we used here. For large $M$, the memory and computational speed bottleneck in an implementation then becomes the large matrix multiplication, ${\bf A}_{\rm train}{\bf A}_{\rm train}^{\top}$. However, there are simple methods that can still enable solution of Eqn.~(\ref{Final}) when  $M$ is too large for this multiplication to be carried out in one calculation.

For example, when solving Equation~(\ref{Final}) by implementation in MATLAB, it is computationally efficient to use the overloaded `$\backslash$' function, which invokes the QR decomposition method.  This approach can be used either for the inverse or pseudo-inverse, but we have found it faster to solve~(\ref{Final}), which requires the inverse rather than the pseudo-inverse.

Well known software packages such as MATLAB (which we used) automatically exploit multiple CPU cores available in most modern PCs to speed up execution of this algorithm using multithreading. Alternative methods like explicitly calculating the pseudo-inverse, or singular value decomposition, are in comparison significantly (sometimes several orders of magnitude) slower.  

When using the linear equation solution method, the main component of training runtime for large hidden-layer sizes becomes the large matrix multiplication required to obtain ${\bf A}_{\rm train}{\bf A}_{\rm train}^{\top}$. There is clearly much potential for speeding up this simple but time-consuming operation, such as by using GPUs or other hardware acceleration methods.

~\\
The above text discusses the standard  single-batch approach. There are also  online and incremental ELM solutions for real-time and streaming operations and large data sets~\cite{Liang.06,Tapson.13,Widrow.13,vanSchaik.14}. The use of singular value decomposition offers some additional insight into network structure and further optimization~\cite{Kasun.13}. Here we describe an iterative method that offers advantages in training where the output weight matrix need not be calculated more than once.

One potential drawback of following the standard ELM method of solving for the output weights using all training data in one batch is the large amount of memory that is potentially required. For example, with the MNIST training set of 60000 images, and a hidden layer size of $M=10000$, the ${\bf A}_{\rm train}$ matrix has $6\times10^8$ elements, which for double precision representations requires approximately $4.5$~GB of RAM. Although this is typically available in modern PCs, the amount of memory required becomes problematic if training data is enhanced by distortions, or if the amount of hidden units needs to be increased significantly.

We have identified  a simple solution to this problem, which is as follows. First, we introduce size $M\times 1$ vectors ${\bf d}_j,~j=1,\dots,K$,  formed from the columns of ${\bf A}_{\rm train}$. Then one of the two key terms in Eqn.~(\ref{Final}) can be expressed as
\begin{align}\label{FinalA}
{\bf A}_{\rm train}{\bf A}_{\rm train}^{\top}= \sum_{j=1}^K {\bf d}_j{\bf d}_j^{\top}.
\end{align}
That is, the matrix that describes correlations between the activations of each hidden unit is just the sum of the outer products of the hidden-unit activations in response to all training data. Similarly, we can simplify the other key term in Eqn.~(\ref{Final}) by introducing size $N\times 1$ vectors ${\bf y}_j,~j=1,\dots K$ to represent the $K$ columns of ${\bf Y}_{\rm label}$ and write
\begin{align}\label{FinalA2}
{\bf Y}_{\rm label}{\bf A}_{\rm train}^{\top}= \sum_{j=1}^K {\bf y}_j{\bf d}_j^{\top}.
\end{align}
In this way, the $M\times M$ matrix ${\bf A}_{\rm train}{\bf A}_{\rm train}^{\top}$ and the $N\times M$ matrix ${\bf Y}_{\rm label}{\bf A}_{\rm train}^{\top}$ can be formed from $K$ training points without need to keep the ${\bf A}_{\rm train}$ matrix in memory, and once these are formed, the least squares solution method applied. The matrix ${\bf A}_{\rm train}{\bf A}_{\rm train}^{\top}$ still requires a large amount of memory ($M=12000$ requires over 1 GB of RAM), but using this method the number of training points can be greatly expanded and incur only a runtime cost. In practice, rather than form the sum from $K$ training points, it is more efficient to form batches of subsets of training points and then form the sums: the size of the batch is determined by the maximum RAM available.

It is important to emphasise that unlike other iterative methods for training ELMs that update the output weights iteratively~\cite{Liang.06,Tapson.13,Widrow.13,vanSchaik.14}, the approach described here only iteratively updates  ${\bf A}_{\rm train}{\bf A}_{\rm train}^{\top}$.

\section{Conclusions}

We have shown here that simple SLFNs are capable of achieving the same accuracy as deep belief networks~\cite{Hinton.06} and convolutional neural networks~\cite{LeCun.98} on one of the canonical benchmark problems in deep learning: image classification.  The most accurate networks we consider here use a combination of several non-iterative learning methods to define a projection from the input space to a hidden layer.  The hidden layer output is then solved simply using least squares regression applied to a single batch of all training data to find the weights for a linear output layer.  If extremely high accuracy is required, the outputs of one or more of these SLFNs can be combined using a simple autoencoder stage.  The maximum accuracy obtained here is comparable with the best published results for the standard MNIST problem, without augmentation of the dataset by preprocessing, warping, noising/denoising or other non-standard modification. The accuracies achieved for the basic SLFN networks are in some cases equal to or higher than those achieved by the best efforts with deep belief networks, for example.  

Moreover, when using the receptive field (RF) method to shape inputs weights, the resulting input weight matrix becomes highly sparse: using the RF algorithm above, close to 90\% of input weights are exactly zero.

We note also that the implementations here were for the most part carried out on standard desktop PCs and required very little computation in comparison with deep networks.  It should be highlighted that we have found significant speed increases for training by avoiding explicit calculation of matrix inverses. Moreover, we have shown that it is possible to circumvent memory difficulties that could arise with large training sets, by iteratively calculating the  matrix ${\bf A}_{\rm train}{\bf A}_{\rm train}^\top$, and then still only computing the output weights once. This method could also be used in streaming applications: the matrix ${\bf A}_{\rm train}{\bf A}_{\rm train}^\top$ could be updated with every training sample, but the output weights only updated periodically. In these ways, we can avoid previously identified potential limitations of the ELM training algorithm, regarding matrix inversion discussed in~\cite{Widrow.13a} (see also~\cite{Widrow.13,Lim.13}).

The principles implemented in the ELM training algorithm, and in particular the use of single-batch least squares regression in a linear output layer, following random projection to nonlinear hidden-units, parallel a principled approach to modelling neurobiological function, known as the {\em neural engineering framework} (NEF)~\cite{Eliasmith}. Recently this framework~\cite{Eliasmith}  was utilized in a very large (2.5 million neuron) model of the functioning brain, known as SPAUN~\cite{Eliasmith.12,Stewart.14a}. The computational and performance advantages we have demonstrated here could potentially boost the performance of the NEF, as well as, of course, the many other applications of neural networks.
 
 Although deep networks and convolutional networks are now standard for hard problems in image and speech processing, their merits were originally argued almost entirely on the basis of their success in classification problems such as MNIST.  The argument was of the form that because no other networks were able to achieve the same accuracy, the unique hierarchy of representation of features offered by deep networks, or the convolutional processing offered by CNNs, must therefore be necessary to achieve these accuracies.  However, If there exists a neural network that does not use a hierarchical representation of features, and which can obtain the same accuracy as one that does, then this argument may be a case of confirmation bias.  We have shown here that results equivalent to those {originally} obtained with deep networks and CNNs {on MNIST} can be obtained with simple single-layer feedforward networks, in which there is only one layer of nonlinear processing; and that these results can be obtained with very quick implementations.  While the intuitive elegance of deep networks is hard to deny, {and the economy of structure of multilayer networks over single layer networks is proven, we would argue that the speed of training and ease of use of ELM-type single layer networks makes them a pragmatic first choice for many real-world machine learning applications.}

\section*{Acknowledgments}

Mark D. McDonnell's contribution was by supported by an Australian Research Fellowship from the Australian Research Council (project number DP1093425). Andr{\'{e}} van Schaik's contribution was supported by Australian Research Council Discovery Project DP140103001.

\section*{References}


\section*{Figures}

\clearpage

\clearpage 

\clearpage

\clearpage

\clearpage

\clearpage

\clearpage

\section*{Tables}

\begin{table}[!ht]
\begin{tabular}{|c|c|c|c|}
\hline
Grouping & Method& Error in testing &	Reference\\
\hline
{Selected Non-ELM} & SLFN, 784-1000-10 &	4.5\% &	\cite{LeCun.98}\\
& Deep Belief Network 	& 1.25\%	& \cite{Hinton.06}\\
& Deep Conv. Net LeNet-5	&0.95\% &	\cite{LeCun.98}\\
& {Deep Conv. Net (dropconnect)} &0.57\% & \cite{Wan.13}\\
& {Deep Conv. Net (stochastic pooling)} &0.47\% & \cite{Zeiler.13}\\
& {Deep Conv. Net (maxout units and dropout)} &0.45\% & \cite{Goodfellow.13}\\
& {Deep Conv. Net (deeply-supervised)} &0.39\% & \cite{Lee.14}\\
\hline
Past ELM & ELM, 784-1000-10 &	6.05\%&	\cite{Tapson.14}\\
& C-ELM	& $\sim$5\%	& \cite{Zhu.15}\\
& CIW-ELM, 784-1000-10   	& 3.55\%	& \cite{Tapson.14}\\
& ELM, 784-7840-10	& 2.75\%	& \cite{Tapson.13}\\
&ELM, 784-unknown-10 & 2.61\%	& \cite{Kasun.13}\\
& CIW-ELM, 784-7000-10	& 1.52\%& 	 \cite{Tapson.14}\\
& ELM+backpropagation, 784-2048-10	& 1.45\%& 	 \cite{Yu.12}\\
&Deep ELM, 784-700-15000-10 & 0.97\%	& \cite{Kasun.13}\\
\hline
ELM \& backprop &  RF-(CIW \& C)-ELM, 784-(2$\times 6400$)-20-500-10 &0.91\% (1.04\%)& 	This report\\
\hline
SLFN ELM  &RF-ELM,784-15000-10	&  1.36\% (1.48\%)& This report\\
& CIW-ELM,784-15000-10	&  1.28\% (1.45\%) & This report\\
&C-ELM, 784-15000-10		&1.26\% (1.33\%)& This report\\	
& RF-CIW-ELM	, 784-15000-10	& 1.03\%(1.10\%)& This report\\
& RF-C-ELM, 784-15000-10		& 0.9\% (0.99\%)& This report\\
& RF-C-ELM, 784-25000-10, Distortions &	0.57\% (0.62\%) & 	This report\\
\hline
DLFN ELM & RF-(CIW\& C)-ELM, 784-(2$\times15000$)-20-500-10&	0.83\% (0.87\%) & 	This report\\
\hline
\end{tabular}
\caption{\small\bf{Comparison of our results on the MNIST data set with published results using other methods. The percentages listed in brackets are the mean error percentage we obtained from 10 independent realisations of each method. The remaining percentage for the results obtained in this report are from the trained ELM out of the 10 repeats with the best results. Values for each trial  shown in Figs~\ref{Fig3} and~\ref{Fig4} demonstrate small spreads either side of the mean value. The abbreviations CIW, C and RF are explained in later sections. Note: SLFN is ``Single-hidden Layer Feedforward Network''; DLFN is ``Dual-hidden Layer Feedforward Network.'' The result described as `Distortions' was obtained by augmenting the training set using affine and elastic distortions, as describe in the main text. `Deep. Conv. Net' is an abbreviation for `Deep Convolutional Network.'}}\label{Table1}
\end{table}%

\clearpage

\end{document}